# Forecasting sales with Bayesian networks: a case study of a supermarket product in the presence of promotions

**M. Hamza [a], M. Abolghasemi [a] and A.O. Alvandi [b]**

*[a] Department of Data Science & AI, Monash University.*
*[b] Department of Human Centred Computing, Monash University*
*Email: mham0030@student.monash.edu, mahdi.abolghasemi@monash.edu and abraham.oshnialvandi@monash.edu*

**Abstract:** Sales forecasting is the prerequisite for a lot of managerial decisions such as production planning, material resource planning and budgeting in the supply chain. Promotions are one of the most important business strategies that are often used to boost sales. While promotions are attractive for generating demand, it is often difficult to forecast demand in their presence. In the past few decades, several quantitative models have been developed to forecast sales including statistical and machine learning models. However, these methods may not be adequate to account for all the internal and external factors that may impact sales. As a result, qualitative models have been adopted along with quantitative methods as consulting experts has been proven to improve forecast accuracy by providing contextual information. Such models are being used extensively to account for factors that can lead to a rapid change in sales, such as during promotions. In this paper, we aim to use Bayesian Networks (BNs) to forecast promotional sales where a combination of factors such as price, type of promotions, and product location impacts sales. We choose to develop a BN model because BN models essentially have the capability to combine various qualitative and quantitative factors with causal forms, making it an attractive tool for sales forecasting during promotions. Also, BNs are graphical tools that allow us to visualize the effect of an observed node on all the other nodes of the network. This can be used to adjust a company's promotional strategy in the context of this case study. We gather sales data for a particular product from a retailer that sells products in Australia. We develop a BN for this product and validate our results by empirical analysis. We show that the BNs are superior in predicting overall average weekly sales and average weekly sales during catalogue promotions to the company's forecasts in the case study. This paper confirms that BNs can be effectively used to forecast sales, especially during promotions. In the end, we provide some research avenues for using BNs in forecasting sales.

## 1. INTRODUCTION AND LITERATURE REVIEW

Companies adopt different modelling and forecasting strategies to implement their plans effectively and reduce their operational costs. Sales are one of the most important factors that companies try to predict as it allows them to effectively allocate resources and plan inventory levels (Wacker & Lummus, 2002). Better forecasts help companies to make better decisions. These decisions may include whether a company wants to have promotions, what kind of advertising will lead to better sales, and other factors such as the location of the product in the store can also have important implications for sales as was studied by Drèze et al. (1994). There are several qualitative and quantitative sales forecasting methods that companies can employ to predict their sales accurately and drive their business successfully. Green et al. (2011) discuss several approaches, both qualitative and quantitative, for sales forecasting. Qualitative methods include Delphi, scenario planning, and judgmental forecasting. Statistical and machine learning techniques are the most available and promising quantitative methods that have been studied massively to forecast sales such as the use of ARIMA regression and Bayesian inference by Stergiou (1991) and Yelland (2010) respectively. Neural networks have also proven to be useful tools in sales forecasting (Yu et al. 2011). More recent studies on self-training deep neural networks have outperformed decision trees and support vector regression in sales predictions in the fashion retail industry (Loureiro et al. 2018).

There are various factors such as price, location of a product, advertisement type and competitors' behaviour that can impact the sales of a product and it is not trivial how these qualitative and quantitative factors can impact the dynamics of sales. While statistical and machine learning models are widely used to forecast product sales, in practice, experts often adjust the output of statistical and machine learning forecasts to project the impact of external factors such as promotion and competitors' behavior. Arora & Taylor (2018) used rule-based forecasting in their predictions which incorporates expert judgement to improve the accuracy of their model when compared to purely quantitative methods. The retailer's sales data used in our research study shows a stark difference between sales during promotions and in the absence of promotions. Such behavior is not surprising as people tend to buy more when there are discounted prices (Jallow & Dastane, 2016), and this also leads to lower sales in periods of normal prices. Promotions can be problematic to models using quantitative methods alone because they depend on contextual information such as the marketing strategy, competitors' behaviour, and other external factors. Therefore, quantitative methods alone may not take into account all the important information and yield less accurate forecasts. Experts have been shown to add value to sales forecasts in the presence of promotions (Trapero et al. 2013).

In this paper, we propose to use Bayesian networks (BNs) as a technique for forecasting sales. BNs is one of the highly effective methods that has found its applications in fields as diverse as prediction of risks of diseases (Arora et al. 2019), forecasting demand of electricity (Bassamzadeh & Ghanem, 2017), energy prediction (Abolghasemi & Alizadeh, 2014), measuring the performance of the supply chain (Abolghasemi et al. 2014), etc. However, BNs have not been used in forecasting sales in the presence of promotions. Since sales forecasting is usually done by relying on historical data, the success of BNs can have huge implications for forecasting sales of products in the absence or lack of historical information such as during promotions. It is expected that by using a case study approach, where historical data is present, to build a BN model, its performance can be compared to the retailer's forecasts. In this work, we use BNs to forecast sales for a particular supermarket product. We will employ data from a popular supermarket that will be used to construct a BN model. We compare the BN generated forecast with the retailer's method to validate our results.

The rest of the paper is organized as follows. In Section 2, we introduce BNs. Section 3.1 discusses our dataset and section 3.2 describes our proposed model. In Section 4, we discuss our empirical results. Finally, Section 5 concludes this study.

## 2. BAYESIAN NETWORKS

BNs are directed, acyclic graphical structures. They consist of nodes that are connected with each other using arcs. A node connected with another node using a directed arc symbolizes a parent-child or cause-effect relationship in the direction of the arc (Heckerman, 1997). In Figure 1, A is the parent of B and C. Similarly, B and C are the parents of D. These relationships should not violate the basic structural principle of BNs. According to this principle mentioned by Nilsson (1998) in his book, any node in the network is conditionally independent of its non-descendants given its parents. For Figure 1, this means that given we know the state that the random variable A is in, the nodes B and C are independent of each other. To be able to conduct inference, we need to feed prior probabilities to the BN. For any parent-child relationship, this means giving conditional probabilities to the children over all possible outcomes of the child and parent. As an example, in Figure 1, P(C|A) needs to be entered for all possible states of A and C. This model then can be used to calculate

Hamza et al., Forecasting sales with Bayesian networks…

posterior probabilities or update beliefs using Bayes theorem if any node becomes observed i.e., the probability of a particular state of that node becomes 1 (Heckerman, 1997).

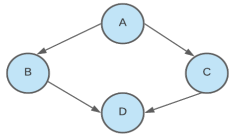

**Figure 1.** A simple BN illustration

### 3. METHODOLOGY

#### 3.1. Data

The data used in this research has been obtained from a retailer and it consists of weekly sales forecasts as well as actual weekly sales along with promotional information such as price, product location in the store (fixture or gondola) and advertisement type (catalogue or in-store) for a product from the end of December 2016 till the end of August 2018. There are a total of 88 observations where various variables can impact sales greatly. After removing one outlier, there are 48 observations associated with promotions (48/87 = 55%) and the remaining 39 observations are without promotions (39/87 = 45%). The promotions data can be further split as in-store promotions (7/87 = 8%) and catalogue promotions (41/87 = 47%). Regardless of the two promotion types, the price discount varies from 31% to 50%. The promotional methods are assumed to be independent. When there is an in-store promotion or no promotion, the product is placed in the gondola (normal aisle in store) whereas catalogue promotions imply that the product is placed in the fixture. As studied by Drèze et al. (1994), this increases the product visibility as compared to normal aisles and contributes significantly to an increase in sales. Both forecast sales data and actual sales data has been split into data with promotions and without promotions and outliers have been removed from each category. Sales data represents the number of units of the product. Promotional sales data is further split into in-store and catalogue promotions. The statistical measures of each category have been described using the notation *N (Mean, SD)* in Table 1.

**Table 1**. Statistical measures of the retailer's data (after removing outliers)

| Data Category | Actual Sales Data | Forecast Sales Data |
|---|---|---|
| Overall | N (163.51, 157.17) | N (159.42, 166.33) |
| No Promotions | N (15.92, 4.4) | N (15.92, 4.84) |
| In-store Promotions | N (90.54, 52.54) | N (100.7, 59.16) |
| Catalogue Promotions | N (312.89, 88) | N (326.98, 105.09) |

#### 3.2. BN Model

The nodes in the BN model represent factors that may directly or indirectly affect sales. An advantage of using BNs is its graphical structure which aids in understanding the influence that different nodes in the network have on each other and on sales. This is because BNs calculate the posterior probabilities for all the nodes in the network after any of the factors becomes observed (i.e., the probability associated with a certain state of that node becomes 1). We consider different nodes for the BN model based on their association in the real-world case study. We then determine the links between them as per case study and literature review. The parameters and association between nodes are estimated empirically. We use 'Promotions', 'Price', 'Product Location', and 'Sales' nodes as follows:

***Promotions***: By counting the number of promotions in the total sales data, the probability associated with promotions and no promotions is 0.55 and 0.45, respectively. However, this node has three possible states. The promotions data is further divided into catalogue and in-store promotions, and their corresponding probabilities are 0.47 and 0.08, respectively based on their occurrence in the data. 'No Promotion' state just means that there is no promotion which has a probability of 0.45.

***Price***: This is a deterministic node and is a child of the 'Promotions' node. This means that given the state of the 'Promotions' node, we will certainly know if this node will be in the 'Normal', 'Discounted Instore' or the 'Discounted Catalogue' state. 'Discounted Instore' and 'Discounted Catalogue' states have been used to differentiate between the promotion types to aid in setting up the mathematical model that will be used in forecasting sales as will be seen in the definition of the 'Sales' node.

*Product Location:* This is a deterministic node and like the 'Price' node, it is a child of 'Promotions'. Given catalogue promotions, the product will always be in the fixture. Whereas for in-store promotions or no promotions, the product will always be in the gondola. Again, 'Gondola InStore' and 'Gondola NP' (no promotions) have been differentiated to help in defining the equation for the 'Sales' node.

*Sales:* This variable has been modelled using an equation node to capture its continuous nature. It is directly connected to 'Price', 'Product Location', and 'Promotions' as these nodes influence sales. However, the weights assigned to these three parent nodes are not equal. From the data, it is seen that given promotions exist, both catalogue and in-store promoted products have similar price discounts. However, the average catalogue promoted products sales (312.9) exceeded the average in-store promoted product sales (90.5) by more than 3.45 times. Without promotions, the average sales are 15.9. Another way to look at this is that the total difference between the lowest category of sales (average sales without promotions) and the highest category (average sales with catalogue promotions) is 312.9 – 15.9 = 297. Since in-store and catalogue promoted products offer similar price discounts only about ((90.5 - 15.9) / 297) = 25% of the variation is attributable to price discounts (both in-store promoted products and non-promotional products are placed in the gondola). 'Price' has been assigned a weight of 25% whereas the remaining 75% has been equally split between 'Product Location' and 'Promotions' as there is no additional information that suggests any other distribution for the remaining weight of 75%. Defining the 'Sales' node requires making appropriate assumptions backed by mathematical explanations to get the best possible results from this model. The roles of 'Price', 'Product Location', and 'Promotions' in the sales equation have been explained in detail below:

The forecast sales data has been used for these nodes. As mentioned in section 3.1, this data was divided into promotional and non-promotional data and the outliers for both the series were removed using the Tukey boxplot definitions of upper and lower fences.

**Price Node:** No promotions imply 'Normal' price. The data for no promotion sales did not follow a normal or lognormal distribution and has been estimated using a triangular distribution (Triangular (9.6, 12, 24)) because we knew the upper and lower limits as well as the mode value from the data. Here 9.6 is the minimum value, 12 is the mode, and 24 is the maximum value. Since 'Price' node has been assigned a weight of 25%, the 'Normal' price state is given by 0.25*Triangular (9.6, 12, 24).

'Catalogue' promotions imply 'Discounted Catalogue' and 'In Store' implies 'Discounted Instore' prices respectively. Although the combination of catalogue and in-store promotion series follows a normal distribution, each of the promotion types separately fails the Shapiro-Wilk normality test. After taking natural log of both of these series, the normality test was again applied and for in-store promotions and catalogue promotions, we retain the null hypothesis that their natural log data is normally distributed at 0.05 and 0.02 significance levels respectively. Given the natural log of a variable is normally distributed, the variable itself can be modelled using a lognormal distribution with mean and standard deviation of the natural log of the data. Hence, we conclude that sales in-store promotions and sales catalogue promotions are lognormally distributed with means (SD) of 4.487 (0.5242) and 5.7466 (0.2889) respectively.

The overall formula for the price node is **Choose ((Price, 0.25*Triangular (9.6, 12, 24), Lognormal (3.1, 0.5242), Lognormal (4.36, 0.2889))** where the choose function selects the triangular distribution output for the 1st state ('Normal' price), Lognormal (3.1, 0.5242) for the 2nd state ('DiscountedInstore' price) and Lognormal (4.36,0.2889) for the 3rd state ('DiscountedCatalogue' price) respectively based on the probabilities of occurrence of these states.

The 25% weight has been used as a scalar multiple of the lognormal distribution in the price node equation. The above is true because given a lognormally distributed variable X with mean μ and standard deviation σ, $X = e^{(\mu+\sigma \cdot Z)}$ where Z is a normally distributed random variable, and $cX = c \cdot e^{(\mu+\sigma \cdot Z)} = e^{\ln(c)} \cdot e^{(\mu+\sigma \cdot Z)} = e^{(\mu+\ln(c)+\sigma \cdot Z)} = e^{(\beta+\sigma \cdot Z)}$, where c is a constant and $\beta = \mu + \ln(c)$ is the new mean. Note, the standard deviation stays the same. The model distributions and parameters (adjusted after scalar multiplication for lognormal distribution) have been summarized in Table 2.

Sales is defined as the addition of the role 'Price', 'Product Location', and 'Promotions' play in its calculation:

**Sales = Choose ((Price, 0.25*Triangular (9.6, 12, 24), Lognormal (3.1, 0.5242), Lognormal (4.36, 0.2889)) + Choose (Promotions, Lognormal (4.766,0.2889), Lognormal (3.507,0.524), 0.375 * Triangular (9.6, 12, 24)) + Choose (Product Location, Lognormal (4.766,0.2889), Lognormal (3.507,0.524), 0.375 * Triangular (9.6, 12, 24))** (1)

This is because we estimated the distributions for the retailer's forecasted sales data for the different categories (i.e., no promotions, in-store promotions, and catalogue promotions). We then assigned weights to the three parent nodes of sales to split their total effect on sales. The BN model then selects one state of each node based on the relative probabilities of these states which is accomplished by the choose function. This will give a numerical value for each node based on the corresponding probability distribution selected in an iteration. Adding the values for each node gives a sales forecast. After averaging these forecasts over 10,000 iterations, the BN model gives us its average weekly sales forecast.

When the BN model is updated, we combine the effects of the 'Price', 'Product Location', and 'Promotions' nodes in the 'Sales' node considering the probabilities of occurrence of the states of these nodes. This is also the reason why our BN model is robust to the selection of weights. The weights for 'Product Location' and 'Promotions' nodes were assumed to be the same (37.5% each) in (1) whereas the price node was assigned a weight of 25% based on the direct effect that it had on sales. Since these weights were more or less arbitrary, a sensitivity analysis was performed to calculate sales using different combinations of weights for 'Product Location' and 'Promotions' nodes such as 30%-45% and 35%-40% of these nodes. It was found that the average weekly sales remained unchanged in each case. However, these weights highlight the relative importance of each node's effect on sales and should be chosen based on their observed effect on sales from data if applicable.

**Table 2.** Distributions and Parameters of the different states in the nodes affecting sales

| Price Node | | Distribution | Parameters |
|---|---|---|---|
| State 1 | Normal | Triangular | Min = 9.6, Mod = 12, Max = 24 |
| State 2 | Discounted Instore | Lognormal | Mean = 3.1, SD = 0.5242 |
| State 3 | Discounted Catalogue | Lognormal | Mean = 4.36, SD = 0.2889 |
| **Promotions Node** | | | |
| State 1 | Catalogue | Lognormal | Mean = 4.766, SD = 0.2889 |
| State 2 | In Store | Lognormal | Mean = 3.507, SD = 0.524 |
| State 3 | No Promotion | Triangular | Min = 9.6, Mod = 12, Max = 24 |
| **Product Location Node** | | | |
| State 1 | Fixture | Lognormal | Mean = 4.766, SD = 0.2889 |
| State 2 | Gondola InStore | Lognormal | Mean = 3.507, SD = 0.524 |
| State 3 | Gondola NP | Triangular | Min = 9.6, Mod = 12, Max = 24 |

The product is only placed in the fixture when there is a catalogue promotion. So, the distributions of 'Promotions' and 'Product Location' nodes are identical because both nodes have the same weightage. The structure of the BN model has been shown below as it has been developed in GeNIe Academic:

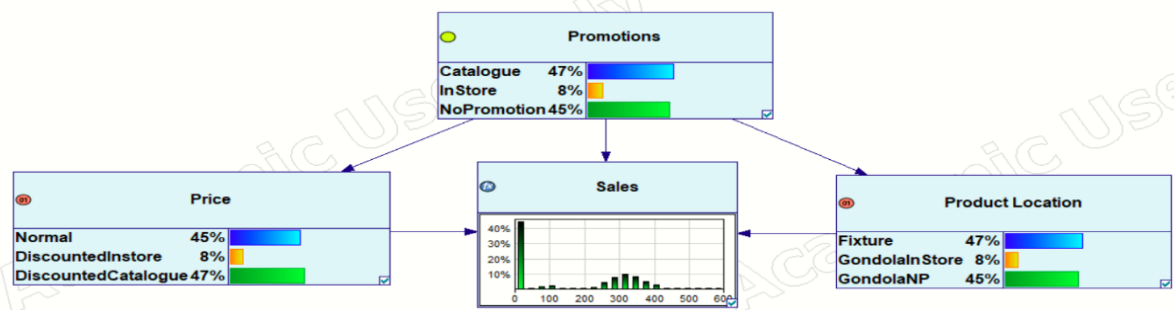

**Figure 2.** BN model for forecasting sales using forecast sales data of a retailer

The above Figure represents the BN model obtained after feeding the prior probabilities. The overall average weekly sales without the effect of any observed variable are noted after updating the network. Then, for each 'NoPromotion', 'Catalogue', and 'InStore' states of the 'Promotions' node, the probability is set to one. The outcome for the 'Sales' node is compared with the retailer's forecast sales data. The results have been summarized in Table 3. Other than studying the effect of observed promotions on sales, the above model is also useful for obtaining the posterior probabilities of the states in 'Promotions' node and the associated nodes of 'Price' and 'Product Location' when a certain level of sales is observed. For example, after updating the model with prior probabilities, we receive evidence in a later period that sales are 175 units, and we would like to find the likelihood of promotions as well as the values of price and product location.

Hamza et al., Forecasting sales with Bayesian networks…

In Figure 3, it can be seen that in-store promotions have the highest likelihood of 67% (the probability of in-store promotion being the reason for achieving Sales = 175 units) which is followed by catalogue promotions at 32% and finally the likelihood of no promotions is extremely low at 1%. This is because the sales volume differs greatly between no promotions and promotions. In the above model, the 'Price' and 'Product Location' nodes are deterministically dependent on 'Promotions' due to the simple relationships present in our dataset. Hence, the posterior probabilities are similar. Such backward analysis can be used by the companies to make strategic decisions such as the right selling price, product marketing strategy, etc. to reach a particular level of sales.

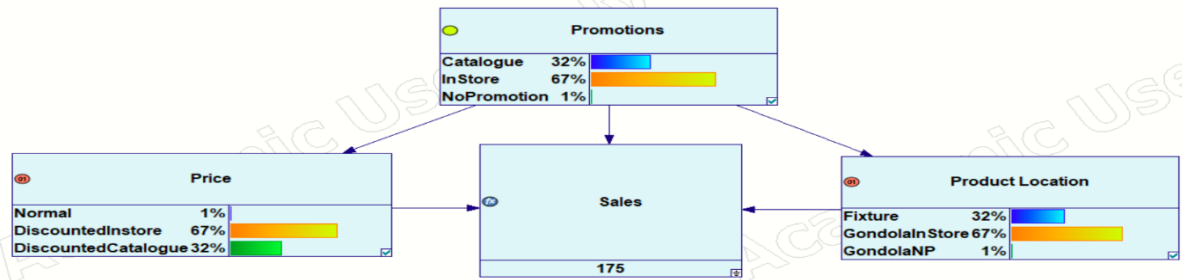

**Figure 3.** Posterior probabilities of 'Price', 'Promotions', and 'Product Location' when Sales = 175

## 4. EMPIRICAL RESULTS AND DISCUSSION

This section presents the results obtained from running the BN model and compares them with the results obtained from the retailer's sales forecasting method. We are not aware of the company's forecasting method. However, we will use their forecasts and calculate overall weekly sales averages and the averages for different promotion types. We compare the forecast accuracy using mean absolute percentage error (MAPE) as it is commonly used to evaluate the performance of forecasting models (Fildes et al. 2009; Abolghasemi et al. 2020). We can calculate MAPE for a single observation by $\frac{abs(f-x)}{x}$, where $x$ is the actual value of the series and, $f$ is the forecast value of the series. Table 3 shows our obtained results in terms of MAPE for various observations.

**Table 3.** MAPE comparison of retailer's sales forecasts with BN sales predictions

| Sales Periods (evidence of Promotion node) | Retailer's forecasts | BN forecasts | Confidence Interval for BN forecasts (at 5% significance level) |
|---|---|---|---|
| Average Weekly Sales (Overall) | 3% | 1% | (162.72, 168.82) |
| Average Weekly Sales No Promotion | 1% | 5% | (15.15, 15.23) |
| Average Weekly Sales In-store Promotions | 11% | 13% | (101.33, 102.67) |
| Average Weekly Sales Catalogue Promotions | 5% | 4% | (325.34, 327.54) |

As shown in Table 3, the average weekly sales include all promotional and non-promotional data. After the BN model is run, it computes the variable of interest for 10,000 iterations and averages the outcome. In this case, our model (1% error) slightly outperforms the retailer's method (3% error). For average sales without promotions, the retailer's method provides very accurate results (1% error) as it is often easier to forecast non-promotional sales. For both average sales with in-store promotions and catalogue promotions, the forecast data by the retailer and BN predictions are quite similar. Overall, the BN predictions are quite competitive when compared with the retail's sales forecasting technique, even outperforming it for average weekly sales (overall) and average weekly sales with catalogue promotions. The 95% confidence intervals for our BN forecasts have been shown in Table 3.

## 5. CONCLUSION

In this paper, we proposed the use of BNs for forecasting sales during promotions. We obtained real-world data from a case study in which a combination of quantitative and qualitative information impacted the sales of a product during promotions. We estimated the distribution of each factor using historical data. We then built a BN model that effectively modelled sales of the product. We tested our model on a particular product and empirically validated the results. Our proposed model outperforms the current industry practice by 2% for overall average weekly sales and 1% for weekly catalogue promoted sales as shown in Table 3.

The model used in this research paper is a rather small model with simple relationships as it captured the data from a case study in which only a few factors were affecting sales greatly with little interference from external factors. However, in several situations, it is expected that models will be more complex and macroeconomic factors that are outside the control of the business will also affect sales. Given the adequate performance of BNs in sales forecasting, this model can be developed with some considerations to represent more complex situations. Using BNs in sales forecasting is particularly useful for products that lack historical data as forecasting of such products is usually done based on qualitative methods such as market surveys or relying on historical data of closely related products which may not be always applicable. Hence, it is expected that BNs will offer a great solution in such situations as expert knowledge can be used to elicit prior probabilities in the absence of historical data and Bayesian inference will be used to make sales predictions.

This is the first attempt to use BNs for sales forecasting in the presence of promotions. While BN models are naturally a great tool for combining qualitative and quantitative information such as during promotions, there are some drawbacks to them that can be considered in future studies. BNs, as proposed in this study, can be labour intensive as experts need to build a model for each product. This may not be efficient when dealing with a large number of products. Automating this process can save a lot of costs and speed up the forecasting process during promotions. However, this is the drawback of several judgmental models that are currently used in industries.

**ACKNOWLEDGEMENT**

The BN model was built using GeNIe Modeler, a product of 'Bayes Fusion, LLC'. http://www.bayesfusion.com/